\documentclass[unnumsec,webpdf,contemporary,large]{Main_manuscript}%

\graphicspath{{Fig/}}

\theoremstyle{thmstyleone}%
\theoremstyle{thmstyletwo}%
\theoremstyle{thmstylethree}%
\usepackage{float}

\usepackage{graphicx}
\usepackage{dcolumn}
\usepackage{bm}
\usepackage{booktabs}
\usepackage{placeins}
\usepackage{tabularx}
\usepackage{array}
\usepackage{rotating}
\usepackage{threeparttable}
\usepackage{makecell}
\usepackage{threeparttable}
\usepackage{longtable}
\usepackage{ragged2e}
\usepackage[utf8]{inputenc}
\usepackage[T1]{fontenc}
\usepackage{mathptmx}
\usepackage[edges]{forest}
\usepackage{etoolbox}
\usepackage[table,dvipsnames]{xcolor}
\usepackage{adjustbox}
\usepackage{academicons}
\usepackage{tikz}
\usetikzlibrary{mindmap,backgrounds}
\usepackage{booktabs}
\usepackage{graphicx} 
\usepackage[table]{xcolor}
\usepackage{array}
\usepackage{xltabular}
\usepackage{ragged2e}

\usepackage{booktabs}
\usepackage{pifont} 
\usepackage{threeparttable} 
\newcommand{\cmark}{\ding{51}} 
\newcommand{\pmark}{$\triangle$} 
\newcommand{\nmark}{\textemdash}

\usepackage{colortbl} 

\renewcommand{\thetable}{S\arabic{table}} 
\renewcommand{\thefigure}{S\arabic{figure}}

\definecolor{flowAccessLabel}{HTML}{BFD8EC} 
\definecolor{flowExecutionLabel}{HTML}{B9DED7} 
\definecolor{tableHeader}{HTML}{D9D9D9} 
\definecolor{flowEvidenceLabel}{HTML}{C9E3BC} 

\definecolor{flowCrosscut}{HTML}{F8EFF1} 
\definecolor{flowCrosscutLabel}{HTML}{E4CCD2}
\definecolor{flowAccess}{HTML}{E6F0F8} 
\definecolor{flowExecution}{HTML}{E8F5F3}
\definecolor{flowEvidence}{HTML}{EDF1E8}
\definecolor{flowTranslation}{HTML}{F5F0E6} 
\definecolor{flowTranslationLabel}{HTML}{DDD0B8}

\definecolor{flowDiscovery}{HTML}{F5EFF9} \definecolor{flowDiscoveryLabel}{HTML}{DCC8E8} 
\definecolor{flowClosure}{HTML}{FDF0E4} \definecolor{flowClosureLabel}{HTML}{F0C9A8}

\definecolor{benchEvidence}{HTML}{EDF1E8} 
\definecolor{benchEvidenceLabel}{HTML}{D1DCC8} 
\definecolor{benchWorkflow}{HTML}{E8F5F3} 
\definecolor{benchWorkflowLabel}{HTML}{B9DED7} 
\definecolor{benchGenomics}{HTML}{E6F0F8}
\definecolor{benchGenomicsLabel}{HTML}{BFD8EC} 
\definecolor{benchSpatial}{HTML}{F0F4E8} 
\definecolor{benchSpatialLabel}{HTML}{D4DFC1} 
\definecolor{benchProtein}{HTML}{FAEEF2} 
\definecolor{benchProteinLabel}{HTML}{EBC8D4} 
\definecolor{benchDrug}{HTML}{F1E7F6} 
\definecolor{benchDrugLabel}{HTML}{DCC8E8} 
\definecolor{benchPathology}{HTML}{FAE8D8} 
\definecolor{benchPathologyLabel}{HTML}{F0C9A8}

\newcommand{\flowlabel}[2]{\rowcolor{#1} \multicolumn{5}{l}{\textbf{\strut #2}} \\ }

\usepackage{booktabs} 

\newcommand{\fevpanelwide}[2]{%
\begin{minipage}[t]{0.485\textwidth} \centering \textbf{(#1)}\\[0.10em] \includegraphics[ width=\linewidth ]{#2} \end{minipage}}

\theoremstyle{thmstyletwo}%
\theoremstyle{thmstylethree}%
\usepackage{float}
\usepackage{graphicx}
\usepackage{dcolumn}
\usepackage{bm}
\usepackage{booktabs}
\usepackage{placeins}
\usepackage{tabularx}
\usepackage{array}
\usepackage{rotating}
\usepackage{threeparttable}
\usepackage{makecell}
\usepackage{threeparttable}
\usepackage{longtable}
\usepackage{ragged2e}
\usepackage[utf8]{inputenc}
\usepackage[T1]{fontenc}
\usepackage{mathptmx}
\usepackage[edges]{forest}
\usepackage{etoolbox}
\usepackage[table,dvipsnames]{xcolor}
\usepackage{adjustbox}
\usepackage{academicons}
\usepackage{tikz}
\usetikzlibrary{mindmap,backgrounds}
\usetikzlibrary{positioning}
\usepackage{booktabs}
\usepackage{array}
\usepackage{xltabular}
\usepackage{ragged2e}

\usepackage[table]{xcolor} 
\usepackage{colortbl} 

\renewcommand{\thetable}{S\arabic{table}} 
\renewcommand{\thefigure}{S\arabic{figure}}

\definecolor{flowAccessLabel}{HTML}{BFD8EC} 
\definecolor{flowExecutionLabel}{HTML}{B9DED7} 
\definecolor{tableHeader}{HTML}{D9D9D9} 
\definecolor{flowEvidenceLabel}{HTML}{C9E3BC} 

\definecolor{flowCrosscut}{HTML}{F8EFF1} 
\definecolor{flowCrosscutLabel}{HTML}{E4CCD2}
\definecolor{flowAccess}{HTML}{E6F0F8} 
\definecolor{flowExecution}{HTML}{E8F5F3}
\definecolor{flowEvidence}{HTML}{EDF1E8}
\definecolor{flowTranslation}{HTML}{F5F0E6} 
\definecolor{flowTranslationLabel}{HTML}{DDD0B8}

\definecolor{flowDiscovery}{HTML}{F5EFF9} \definecolor{flowDiscoveryLabel}{HTML}{DCC8E8} 
\definecolor{flowClosure}{HTML}{FDF0E4} \definecolor{flowClosureLabel}{HTML}{F0C9A8}

\definecolor{benchEvidence}{HTML}{EDF1E8} 
\definecolor{benchEvidenceLabel}{HTML}{D1DCC8} 
\definecolor{benchWorkflow}{HTML}{E8F5F3} 
\definecolor{benchWorkflowLabel}{HTML}{B9DED7} 
\definecolor{benchGenomics}{HTML}{E6F0F8}
\definecolor{benchGenomicsLabel}{HTML}{BFD8EC} 
\definecolor{benchSpatial}{HTML}{F0F4E8} 
\definecolor{benchSpatialLabel}{HTML}{D4DFC1} 
\definecolor{benchProtein}{HTML}{FAEEF2} 
\definecolor{benchProteinLabel}{HTML}{EBC8D4} 
\definecolor{benchDrug}{HTML}{F1E7F6} 
\definecolor{benchDrugLabel}{HTML}{DCC8E8} 
\definecolor{benchPathology}{HTML}{FAE8D8} 
\definecolor{benchPathologyLabel}{HTML}{F0C9A8}

\usepackage{booktabs} 

\usetikzlibrary{arrows.meta}

\begin{document}


\journaltitle{Agentic Bioinformatics through Function, Evidence, and Validation}
\copyrightyear{2026}
\pubyear{2026}

\firstpage{1}


\title[Agentic Bioinformatics and the FEV Framework]{Evaluating Agentic Bioinformatics through Function, Evidence, and Validation}

\author[1]{Phuc Pham} 
\author[1,$\ast$]{Truong Son Hy} 

\address[1]{
\orgdiv{Department of Computer Science}, \orgname{The University of Alabama at Birmingham},  \country{USA}}

\corresp[$\ast$]{Corresponding author. \href{email:thy@uab.edu}{thy@uab.edu}}

\received{Date}{0}{Year}
\revised{Date}{0}{Year}
\accepted{Date}{0}{Year}



\abstract{Large language model agents increasingly plan, execute, and interpret biological analyses, yet fluent responses, successful tool calls, and benchmark performance alone do not establish scientific credibility. Existing reviews primarily organize biological agents by application, architecture, and agentic capability, but do not jointly operationalize the accountability of agent-generated workflows. We address this gap by treating the inspectable workflow trajectory, rather than architecture or final output alone, as the primary unit of analysis. We introduce the Function--Evidence--Validation (FEV) framework, which separates demonstrated workflow operations, traceable support for actions and claims, and use-case-specific validation. Using FEV, we map 109 agentic or agent-adjacent systems and 28 benchmark or evaluation resources, representing 128 unique publications across genomics, single-cell and spatial omics, protein science, drug discovery, computational pathology, and general bioinformatics automation. Across domains, planning and tool-mediated execution have advanced more rapidly than replayability, provenance, robust scientific assessment, external validation, and prospective empirical testing. We therefore argue that agentic bioinformatics should be assessed through workflow correctness rather than final-answer correctness alone. FEV provides a practical basis for comparing systems and designing transparent, auditable, and scientifically accountable bioinformatics workflows.}


\keywords[Abbreviations]{ AI, artificial intelligence; FEV, Function--Evidence--Validation; LLM, large language model; scRNA-seq, single-cell RNA sequencing }




\maketitle


\begin{figure*}[ht]
    \centering
     \includegraphics[width=1\linewidth]{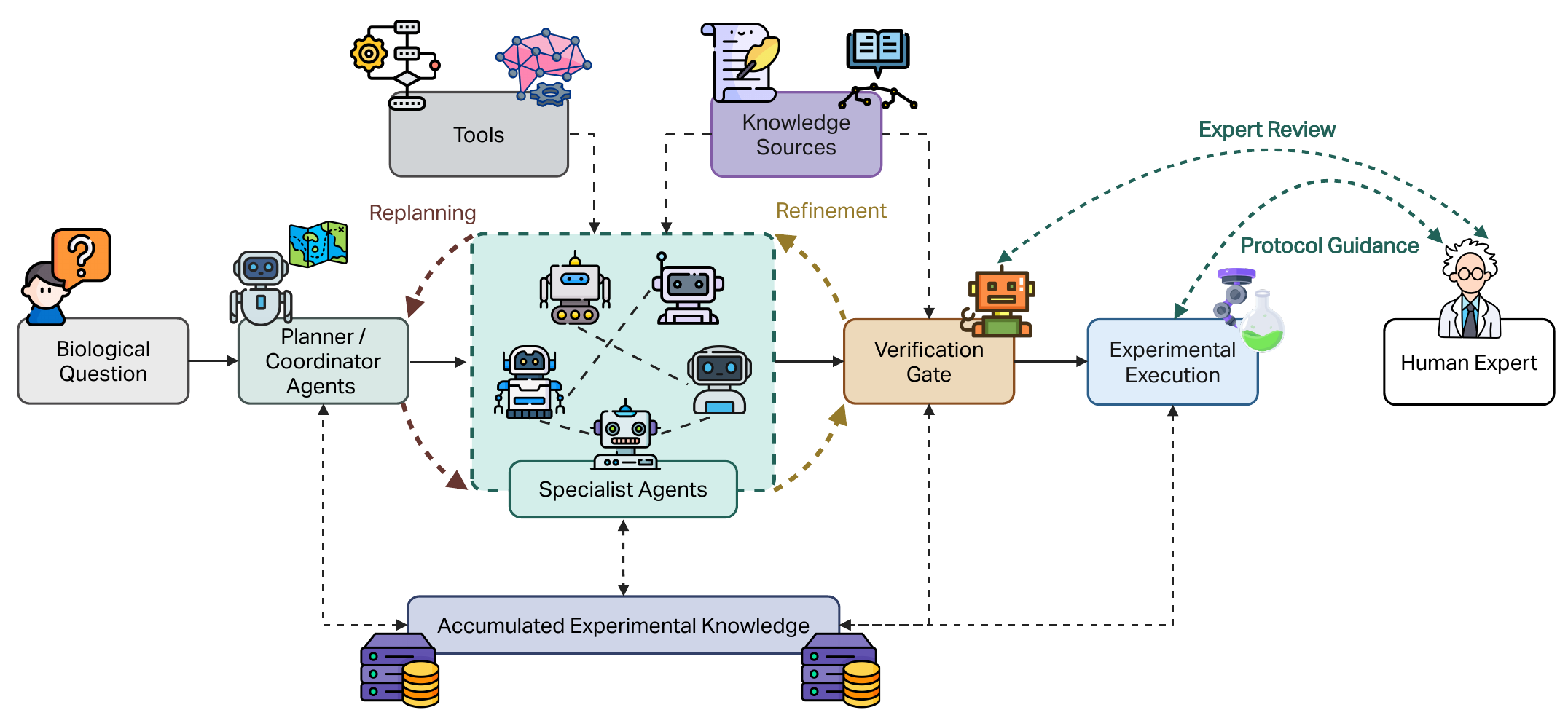}
    \caption{\textbf{Conceptual overview of agentic bioinformatics.} High-level biological questions and/or input data are translated by planner or coordinator agents into coordinated actions across specialist agents, external tools, and knowledge sources. Intermediate outputs pass through a verification gate informed by accumulated experimental knowledge or memory and expert review, enabling iterative replanning and refinement. Depending on workflow scope and the assurance established for the evaluated use case, outputs may support computational interpretation, protocol guidance, or empirically grounded execution.}
    \label{fig:architecture}
\end{figure*}

\section{Introduction}
\label{sec:introduction}

\begin{table*}[!t] \centering \caption{\textbf{Functional distinction among bioinformatics copilots, workflow engines, and agentic bioinformatics layers.} The categories describe workflow roles and are not mutually exclusive.} \label{tab:copilot_engine_agentic_layer} \setlength{\tabcolsep}{3pt} \renewcommand{\arraystretch}{1.20} \begin{tabular}{P{2.5cm} P{4.3cm} P{4.4cm} P{6.3cm}} \toprule \rowcolor{gray!30} \textbf{Dimension} & \textbf{Bioinformatics copilot} & \textbf{Workflow engine} & \textbf{Agentic bioinformatics layer} \\ \midrule \rowcolor{gray!10} \textbf{Primary role} & Assists user-led analysis. & Executes predefined workflows. & Adaptively controls scientific workflows. \\ \textbf{Input} & Prompts, documents, or code. & Data, scripts, parameters, and environments. & Biological objectives, data, evidence, constraints, and tools. \\ \rowcolor{gray!10} \textbf{Output} & Suggestions, explanations, or code snippets. & Outputs, logs, reports, and intermediate files. & Plans, tool calls, artifacts, evidence records, and interpretations. \\ \textbf{Planning} & Primarily user-directed. & Specified in advance. & Goal-directed, multistep, and revisable. \\ \rowcolor{gray!10} \textbf{Tool execution} & Optional or user-mediated. & Central but predefined. & Adaptive and responsive to workflow state. \\ \textbf{Workflow state} & Mostly conversational and transient. & Tracks predefined job execution. & Tracks results, assumptions, failures, and pending actions. \\ \rowcolor{gray!10} \textbf{Traceability} & Usually limited. & Strong execution provenance. & Computational traces and evidentiary provenance. \\ \textbf{Verification} & Primarily performed by the user. & Checks execution and predefined conditions. & Assesses outputs and escalates when evidence is insufficient. \\ \rowcolor{gray!10} \textbf{Typical failure} & Plausible but unsupported advice. & Correct execution of an inappropriate workflow. & Incorrect planning, evidence use, or insufficient validation. \\ \bottomrule \end{tabular} \end{table*}

\subsection{Motivation, Definition, and Contributions} 

Bioinformatics increasingly relies on interdependent workflows linking biological questions to data acquisition, quality control, preprocessing, statistical analysis, visualization, and interpretation. Reliable automation therefore requires more than successful software execution. It also depends on appropriate data and metadata, defensible analytical assumptions, preserved parameters and intermediate artifacts, and conclusions supported by traceable evidence \cite{cohen2017scientific,gao2024empowering,trustworthymultiomics}. 

Large language models (LLMs) provide natural-language interfaces for literature retrieval, code generation, database interaction, and scientific interpretation \cite{yang2025rise,zhou2025large,wang2025large}. Agentic systems extend these capabilities through multistep planning, tool-mediated execution, workflow-state management, critique, repair, and adaptive decision-making. Recent systems span multi-omics analysis, single-cell workflows, protein design, and experimentally evaluated biomolecular discovery \cite{zhou2024ai,xiao2026cellagent,ghafarollahi2024protagents,swanson2025virtual}. We use \emph{agentic bioinformatics} to describe systems that adaptively coordinate data, evidence, models, tools, and analytical actions across a biological workflow. 

Operational capability, however, does not establish scientific reliability. A system may execute an inappropriate analysis, retrieve authoritative but irrelevant evidence, preserve a replayable computation supporting an unjustified interpretation, or experimentally test only a restricted part of its principal claim. Scientific credibility therefore depends on the \emph{inspectable workflow trajectory}: the objective, data and metadata, analytical decisions, tools and parameters, intermediate artifacts, failures and repairs, supporting evidence, and evaluation procedures connecting a biological question to its reported conclusion \cite{lin2025bridging,zhou2025streamline,qi2026artificial}. 

This motivates a shift from \emph{final-answer correctness} to \emph{workflow correctness}. The central question is not simply whether an agent produces a plausible or benchmark-matching answer, but whether the process supporting that answer is appropriate, executable, traceable, replayable, and scientifically defensible. 

Recent reviews have organized biological agents by application domain, architecture, agent capability, interaction mode, evaluation strategy, and resource integration \cite{zhou2025streamline,yang2025rise,qi2026artificial,dip2026large,branda2026next}. They have also identified evidence grounding, reproducibility, robustness, and validation as important concerns. The remaining gap is therefore not the absence of these topics, but the lack of a common framework that jointly operationalizes them at the level of individual evaluated workflows. 

Architecture, tool access, or final performance alone cannot determine what operations a system performed, what evidence supported its actions and claims, or what assurance its evaluation established. Systems with similar architectures may provide substantially different forms of assurance, whereas systems with different architectures may expose comparable workflow traces, evidence records, and validation procedures. We therefore treat the inspectable workflow trajectory, rather than the architecture or final answer alone, as the primary unit of analysis. Table~\ref{tab:review_positioning} places this perspective relative to recent reviews.

\begin{table*}[t] \centering \begin{threeparttable} \caption{ Feature-based positioning of recent reviews of biological and bioinformatics agents. Symbols indicate whether each feature is systematically operationalized (\cmark), explicitly discussed but not used as a principal system-level dimension (\pmark), or not separately operationalized (\nmark). The comparison concerns analytical design rather than overall review quality. } \label{tab:review_positioning} \scriptsize \setlength{\tabcolsep}{3.7pt} \renewcommand{\arraystretch}{1.22} \begin{tabular}{ p{3.10cm} >{\centering\arraybackslash}p{1.08cm} >{\centering\arraybackslash}p{1.08cm} >{\centering\arraybackslash}p{1.08cm} >{\centering\arraybackslash}p{1.10cm} >{\centering\arraybackslash}p{1.10cm} >{\centering\arraybackslash}p{1.10cm} >{\centering\arraybackslash}p{1.10cm} >{\centering\arraybackslash}p{1.18cm} >{\centering\arraybackslash}p{1.15cm} } \toprule \textbf{Review} & \rotatebox{90}{\textbf{Application coverage}} & \rotatebox{90}{\textbf{Architecture taxonomy}} & \rotatebox{90}{\textbf{Capability taxonomy}} & \rotatebox{90}{\textbf{Trajectory as unit}} & \rotatebox{90}{\textbf{Explicit evidence profile}} & \rotatebox{90}{\textbf{Replayability gate}} & \rotatebox{90}{\textbf{Cumulative validation}} & \rotatebox{90}{\textbf{Evidence--testing distinction}} & \rotatebox{90}{\textbf{Standardized system coding}} \\ \midrule Zhou et al. \cite{zhou2025streamline} & \cmark & \cmark & \cmark & \pmark & \pmark & \nmark & \nmark & \pmark & \pmark \\ Yang et al. \cite{yang2025rise} & \cmark & \cmark & \cmark & \pmark & \pmark & \nmark & \nmark & \pmark & \pmark \\ Qi et al. \cite{qi2026artificial} & \cmark & \cmark & \cmark & \cmark & \pmark & \pmark & \pmark & \pmark & \cmark \\ Dip et al. \cite{dip2026large} & \cmark & \cmark & \cmark & \cmark & \pmark & \pmark & \pmark & \pmark & \cmark \\ Branda et al. \cite{branda2026next} & \cmark & \cmark & \cmark & \cmark & \cmark & \pmark & \pmark & \pmark & \pmark \\ \textbf{Present FEV review} & \cmark & \pmark & \cmark & \cmark & \cmark & \cmark & \cmark & \cmark & \cmark \\ \bottomrule \end{tabular} \begin{tablenotes}[flushleft] \footnotesize \item ``Trajectory as unit'' indicates that the inspectable sequence from a scientific objective to an evidence-supported conclusion is the primary unit of analysis. \item ``Replayability gate'' separates demonstrated execution from computation reported with sufficient inputs, parameters, dependencies, artifacts, and traces for reconstruction or replay. ``Cumulative validation'' requires each higher stage to satisfy the requirements of lower stages. \item ``Evidence--testing distinction'' separates previously available empirical observations used as evidence from prospective testing of a workflow-generated output and from closed-loop empirical refinement. \item A dash does not imply complete absence of a topic; it indicates that the feature is not defined and applied as a distinct analytical component. 
\end{tablenotes} \end{threeparttable} \end{table*}

To address this gap, we introduce the \emph{Function--Evidence--Validation} (FEV) framework as an accountability-oriented complement to existing application and architectural taxonomies. Rather than asking only what an agent contains or which task it completes, FEV asks what the reported workflow demonstrates and what scientific conclusions that the workflow can support. 

\emph{Function} records the workflow operations that a system demonstrably performs, including planning, coordination, tool-mediated execution, workflow-state and trace maintenance, repair, and verification or escalation. 

\emph{Evidence} records traceable literature, structured biological knowledge, biological measurements and metadata, software and statistical outputs, scientific-model outputs, and experimental or clinical observations supporting workflow actions and claims. 

\emph{Validation} records the assurance established for the evaluated workflow and its claims. These dimensions are complementary but non-interchangeable. Broader Function does not imply stronger Evidence, and access to stronger Evidence does not establish deeper Validation. For example, a published experimental observation used by a workflow contributes to its Evidence profile, whereas prospective testing of a workflow-generated output contributes to Validation. Similarly, benchmark performance does not establish replayability without identifiable inputs, parameters, dependencies, intermediate artifacts, and execution traces. A complete FEV description combines a Function profile, an Evidence profile, and a use-case-specific reported Validation stage. Function and Evidence describe different workflow properties rather than ordinal levels of maturity. Validation comprises five cumulative assurance stages: illustrative output (\textbf{V0}), demonstrated execution (\textbf{V1}), replayable computation (\textbf{V2}), scientifically evaluated computation (\textbf{V3}), and prospective empirical evaluation (\textbf{V4}). Each stage applies to a specific evaluated use case rather than an entire platform. Separate qualifiers preserve the reported forms of assessment: benchmark or baseline comparison, protocol-based human or expert assessment, statistical or uncertainty analysis, robustness or failure analysis, external or independent validation, prospective empirical testing, and closed-loop empirical refinement. The stage and qualifiers reported should therefore be interpreted together with the corresponding Function and Evidence profiles. Using FEV, we construct a structured cross-domain evidence map comprising 109 agentic or agent-adjacent system entries and 28 benchmark or evaluation resources. Nine publications contribute to both categories, yielding 128 unique publications after accounting for overlap. The map spans genomics and gene editing, single-cell and spatial omics, proteomics and protein design, drug discovery, computational pathology and imaging, and general bioinformatics workflow automation. It compares the reported workflow capabilities and forms of scientific assurance rather than ranking systems, platforms, or biological fields. Our contributions are fourfold: 

\begin{itemize} 

\item We define agentic bioinformatics as an adaptive workflow-control layer and establish the inspectable workflow trajectory, rather than architecture or final output alone, as the primary unit of analysis. 
\item We introduce FEV to separate demonstrated workflow \emph{Function}, traceable \emph{Evidence}, and use-case-specific \emph{Validation}, thereby distinguishing operational breadth from scientific assurance. \item We construct a cross-domain evidence map of 109 system entries and 28 benchmark or evaluation resources, representing 128 unique publications after accounting for overlap. 

\item We derive a workflow-correctness agenda spanning analytical appropriateness, execution, replayability, provenance, evidentiary traceability, uncertainty, robustness, external validation, expert escalation, and prospective empirical testing. 
\end{itemize} 

Together, these contributions shift the evaluation of agentic bioinformatics from asking only \emph{what an agent can do} to asking \emph{what scientific conclusions its reported workflow can support}.

\section{From Bioinformatics Copilots to Agentic Bioinformatics} \label{sec:background} 

Large language models increasingly provide natural-language access to biological data, analytical software, databases, and scientific literature \cite{lin2025bridging,dip2026large}. Many applications function as \emph{bioinformatics copilots}, assisting user-directed analysis through scripting, visualization, retrieval, and interpretation while leaving the biological objective, analytical design, tool selection, and execution primarily under human control \cite{wang2024bioinformatics1,wang2024bioinformatics2,wang2024code}. Although such systems can reduce computational barriers, they generally do not maintain or adapt a persistent scientific workflow. 

Conventional workflow engines address a complementary problem. Frameworks such as Nextflow and nf-core provide dependency management, environment specification, scalable execution, and provenance for predefined pipelines \cite{di2017nextflow,langer2025empowering}. They improve procedural reproducibility, but generally do not determine whether a workflow, method, or parameterization is appropriate for the biological objective. A workflow engine can therefore execute an analytically inappropriate pipeline correctly. 

An \emph{agentic bioinformatics layer} adds adaptive workflow control. It interprets biological objectives, decomposes them into computational tasks, selects and invokes tools, maintains the workflow state, inspects intermediate outputs, responds to failures, and determines when additional evidence, revision, or review is required \cite{su2025biomaster,zhou2025streamline,dip2026large}. Rather than replacing workflow engines, this layer can use them as execution substrates while providing planning, evidence integration, and workflow-level decision-making. Table~\ref{tab:copilot_engine_agentic_layer} summarizes these distinctions. The categories are functional rather than mutually exclusive: a platform may combine conversational assistance, predefined workflow execution, and agentic control across different tasks. 

For this review, a full agentic bioinformatics layer demonstrates three minimum capabilities: 

\begin{enumerate} 

\item \textbf{Goal-directed multistep control:} translation of a biological objective into an ordered and revisable computational plan; 

\item \textbf{Stateful tool-mediated execution:} invocation or coordination of analytical tools while retaining sufficient workflow state to interpret intermediate outputs and respond to failures; and 

\item \textbf{Inspectable scientific support:} preservation of explicit evidence, execution traces, verification outputs, or escalation records beyond single-turn response generation. 
\end{enumerate} 

Systems demonstrating only a subset of these capabilities are retained as \emph{agent-adjacent} when they provide relevant evidence for the emerging workflow paradigm. This boundary captures the transition from interactive assistance to adaptive workflow control without assuming that planning, tool access, or multi-agent architecture alone establishes scientific reliability. Unless otherwise stated, aggregate analyses include both full agentic and agent-adjacent system entries retained in the evidence map.

\section{A Three-Dimensional Framework for Scientific Workflow Accountability} \label{sec:taxonomy}

\begin{figure*}[ht]
    \centering
     \includegraphics[width=1\linewidth]{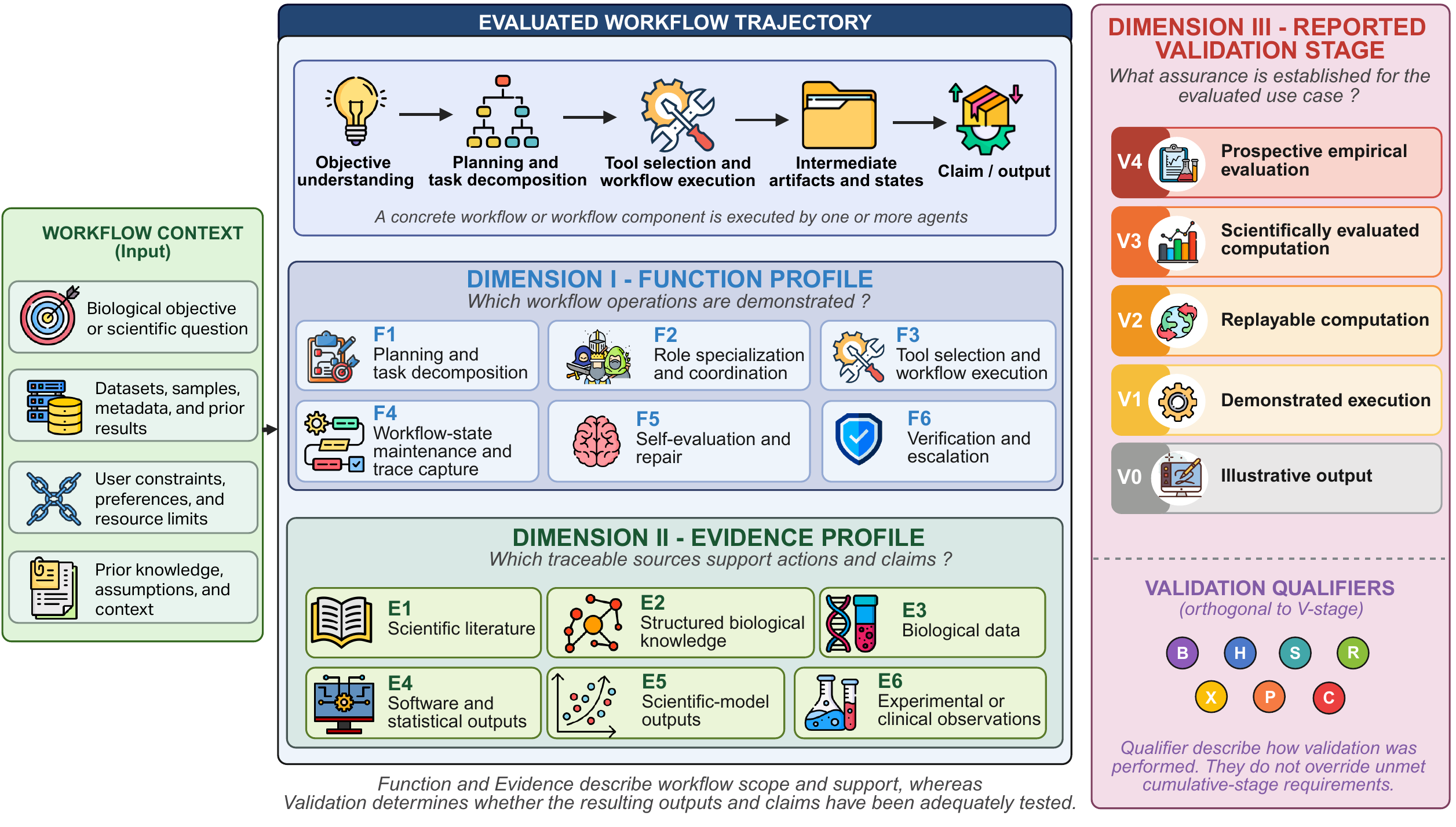}
\caption{ \textbf{Integrated overview of the Function--Evidence--Validation (FEV) framework for agentic bioinformatics.} A biological objective, associated data and metadata, user constraints, and prior context define the evaluated workflow trajectory. \textit{Function} (F1--F6) describes the workflow operations demonstrated by the system, including planning, coordination, tool-mediated execution, workflow-state and trace maintenance, self-evaluation and repair, and verification or escalation. \textit{Evidence} (E1--E6) identifies the traceable sources supporting workflow actions and scientific claims, including literature, structured biological knowledge, biological measurements and metadata, software and statistical outputs, scientific-model outputs, and experimental or clinical observations. \textit{Validation} reports the cumulative assurance established for a specific evaluated use case, progressing from illustrative output (V0) and demonstrated execution (V1) to replayable computation (V2), scientifically evaluated computation (V3), and prospective empirical evaluation (V4). Validation qualifiers specify the reported forms of assessment: benchmark or baseline comparison (B), human or expert assessment (H), statistical or uncertainty analysis (S), robustness or failure analysis (R), external or independent validation (X), prospective empirical testing (P), and closed-loop empirical refinement (C). Function and Evidence are complementary profiles, whereas the reported V-stage is use-case-specific and does not constitute a global ranking of the system or platform. } \label{fig:fev_taxonomy}
\end{figure*}

The Function--Evidence--Validation (FEV) framework characterizes the scientific accountability of agentic bioinformatics workflows through three complementary but non-interchangeable dimensions. \textbf{Function} describes the workflow operations demonstrated by a system. The \textbf{evidence} describes the traceable sources used to inform these operations and support the resulting interpretations. \textbf{Validation} describes the procedures used to test whether workflow outputs and scientific claims are reliable. Figure~\ref{fig:fev_taxonomy} provides an integrated overview of these three dimensions. The dimensions represent different analytical views of the same workflow trajectory rather than sequential processing stages. Greater functional capability does not imply stronger evidence grounding, and access to stronger evidence does not by itself constitute deeper validation. A complete FEV description therefore combines a Function profile, an Evidence profile, and a use-case-specific reported Validation stage.

\subsection{Function: Demonstrated Workflow Operations} \label{subsec:functional-taxonomy} 

The Function dimension characterizes the workflow operations that an agentic system demonstrably performs. This dimension is architecture-agnostic: a system may be implemented as a single agent, a multi-agent team, a tool-augmented language model, or a hybrid workflow controller. Its Function profile is determined by observable workflow behavior rather than by model family, agent count, or architectural label. 

Across the reviewed systems, six recurring dimensions emerge: planning and task decomposition; role specialization and coordination; tool selection and workflow execution; workflow-state maintenance and trace capture; self-evaluation and repair; and verification and escalation. Together, these dimensions describe how systems move from conversational assistance toward adaptive workflow control. They are descriptive and non-ordinal: broader Function profiles do not necessarily imply stronger evidence grounding, replayability, or scientific Validation. Table~\ref{tab:functional_taxonomy} operationalizes each dimension through an observable question, representative mechanisms, and its contribution to the workflow. 

\begin{table*}[!t] \centering \footnotesize \caption{\textbf{Function dimension of the FEV framework.} The table summarizes observable workflow operations that an agentic bioinformatics system may demonstrate; these capabilities do not constitute a maturity score.} \label{tab:functional_taxonomy} \setlength{\tabcolsep}{3pt} \renewcommand{\arraystretch}{1.20} \begin{tabular}{P{3.4cm} P{4.4cm} P{5.1cm} P{4.8cm}} \toprule \rowcolor{gray!30} \textbf{Functional capability} & \textbf{Operational question} & \textbf{Typical mechanisms} & \textbf{Workflow contribution} \\ \midrule \rowcolor{gray!10} \textbf{Planning and task decomposition} & Can the system translate a biological objective into an ordered and revisable plan? & Goal decomposition, dependency-aware planning, step ordering, decision points, and replanning. & Converts biological intent into computational subtasks. \\ \textbf{Role specialization and coordination} & Can the system coordinate specialized agents, modules, tools, or human contributors? & Planner--executor--reviewer roles, specialist modules, shared state, task allocation, and explicit handoffs. & Coordinates biological reasoning, execution, interpretation, and review. \\ \rowcolor{gray!10} \textbf{Tool selection and workflow execution} & Can the system select, parameterize, invoke, and monitor appropriate tools or databases? & API calls, code execution, database queries, parameter selection, output parsing, and workflow-engine invocation. & Produces executable analyses and intermediate artifacts. \\ \textbf{Workflow-state maintenance and trace capture} & Can the system retain workflow context and record what occurred? & Persistent task state, intermediate results, tool logs, parameters, versions, dataset identifiers, and artifact lineage. & Supports continuity, inspection, replay, and audit. \\ \rowcolor{gray!10} \textbf{Self-evaluation and repair} & Can the system detect failures or weak outputs and revise its actions? & Execution monitoring, error diagnosis, critic feedback, consistency checks, code repair, and replanning. & Limits error propagation across multistep analyses. \\ \textbf{Verification and escalation} & Can the system assess output sufficiency and determine when additional review is required? & Tool checks, analytical diagnostics, plausibility checks, uncertainty flags, refusal, clarification requests, and approval gates. & Supports verification and appropriate escalation. \\ \bottomrule \end{tabular} \end{table*} 

\paragraph{Planning, coordination, and execution.} 

Planning requires more than generating a plausible list of analytical steps. A planning-capable system translates a biological objective into an ordered workflow, identifies inputs and dependencies, and revises its plan when information is missing, execution fails, or intermediate results make the initial strategy unsuitable. This distinction is particularly important in omics workflows, where choices that involve quality control, normalization, sample grouping, feature selection, and batch correction affect downstream inference \cite{zhou2024ai,xiao2026cellagent,luecken2019current, cuevas2024data}. 

Role specialization records whether the responsibilities for planning, execution, biological interpretation, critique, or review are connected through explicit task allocation, shared state, and interpretable handoffs \cite{dip2026large,bu2026empowering,xiao2026cellagent}. The presence of multiple agents alone is insufficient to establish coordination. 

Similarly, access to a tool library does not establish tool-mediated execution. The system must select an appropriate tool, construct valid inputs, supply parameters, monitor execution, parse its outputs, and pass the resulting artifacts to subsequent workflow steps. Successful execution demonstrates operational capability, but the suitability of the method and the resulting interpretation must be assessed separately through Evidence and Validation \cite{huang2025biomni,ma2026toolsgenie,mitchener2025bixbench, guo2026promptbio}. 

\paragraph{Workflow state and trace capture.} 

Long-running analyses require more than conversational memory. Workflow-state maintenance allows a system to retain completed steps, intermediate outputs, active assumptions, failures, and unresolved decisions. Trace capture instead exposes what occurred through records such as tool calls, parameters, software versions, dataset identifiers, intermediate artifacts, and output lineage. 

The two functions are related, but not equivalent: a system may maintain a rich internal state while exposing little usable provenance, whereas a conventional workflow engine may preserve strong execution traces without agent-like memory \cite{du2026memory,hu2025memory,kanwal2017investigating, khan2019sharing}. Function records whether state and traces are captured; whether the reported records are sufficient to reconstruct or replay the evaluated computation is determined under Validation. 

\paragraph{Self-evaluation, repair, and verification.} 

Self-evaluation and repair require a system to detect an execution failure, weak intermediate result, or unsupported interpretation and consequently modify its plan, code, tool use, candidate selection, or interpretation. Relevant mechanisms include execution monitoring, error diagnosis, critic feedback, tool substitution, code correction, and replanning \cite{yao2023react,shinn2023reflexion,madaan2023self}. The presence of a critic or reflection component alone is insufficient, unless its assessment changes the workflow or output. 

Verification and escalation address a different question: is an intermediate artifact, a final output, or a proposed claim sufficiently supported to proceed. Verification may involve schema and file checks, analytical diagnostics, biological plausibility assessment, evidence-coverage checks, uncertainty signaling, clarification requests, refusal, or approval gates. Its result may allow an output to proceed, flag or defer it, trigger an additional review, or lead to escalation when available evidence is insufficient \cite{mohammadi2025evaluation,ma2024agentboard}. Verification as a Function records that such an assessment occurs; whether the procedure is scientifically adequate or improves reliability belongs to the Validation dimension. 

Together, the six dimensions form a profile of what a system can plan, coordinate, execute, retain, repair, and verify. The breadth of the functions describes the scope of the workflow rather than autonomy, quality, or trustworthiness. A broadly capable system may rely on weak or poorly traceable evidence, whereas a narrowly scoped system may provide strong Evidence and Validation for a specific task. Function must therefore be interpreted jointly with the sources supporting workflow actions and the assurance established for the evaluated use case.

\subsection{Evidence: Traceable Support for Workflow Actions and Claims} \label{subsec:evidence-taxonomy} 

The Evidence dimension characterizes the sources used to inform workflow actions and support scientific interpretations. It is evaluated separately from Function because planning, tool use, and report generation may still rely on generic model priors, untraceable retrieval, inappropriate data, or unsupported intermediate outputs. 

Across the reviewed systems, six recurring evidence categories emerge: scientific literature; structured biological knowledge; biological measurements and metadata; software and statistical outputs; scientific-model outputs; and experimental or clinical observations. These categories form a descriptive and non-ordinal Evidence profile. Their relevance depends on the workflow and claim: literature may support method selection or interpretation, biological measurements may support dataset-specific inference, and empirical observations may provide context, corroboration, or feedback. 

The quality of the evidence depends not only on source type but also on traceability, specificity, independence, reliability, and analytical role. An identifiable source may still be outdated, methodologically weak, contextually irrelevant, or insufficiently specific to the evaluated claim. Evidence objects should therefore remain linked to their origins and versions and be distinguished according to whether they serve as inputs, intermediate artifacts, contextual support, corroboration, or Validation targets. Table~\ref{tab:evidence_taxonomy} summarizes these categories, their workflow roles, traceability requirements, and principal limitations.

\begin{table*}[!t] \centering \footnotesize \caption{\textbf{Evidence dimension of the FEV framework.} Evidence categories describe the sources supporting workflow actions and scientific claims; they do not constitute an ordinal ranking.} \label{tab:evidence_taxonomy} \setlength{\tabcolsep}{3pt} \renewcommand{\arraystretch}{1.20} \begin{tabular}{P{3.3cm} P{4.8cm} P{4.6cm} P{5.0cm}} \toprule \rowcolor{gray!30} \textbf{Evidence category} & \textbf{Workflow role} & \textbf{Traceability requirement} & \textbf{Principal limitations} \\ \midrule \rowcolor{gray!10} \textbf{Scientific literature} & Supports method selection, hypothesis generation, and interpretation. & Citations, persistent identifiers, retrieved passages, publication version, and retrieval date. & Incomplete coverage, publication bias, outdated findings, and weak dataset-specific support. \\ \textbf{Structured biological knowledge} & Supports entity normalization, annotation, pathway reasoning, and known biological relationships. & Database name and version, identifiers, mappings, query records, and evidence codes. & Curation lag, incomplete coverage, identifier mismatch, and context-insensitive relationships. \\ \rowcolor{gray!10} \textbf{Biological measurements and metadata} & Supports dataset-specific analysis and empirical inference. & Dataset accession, sample metadata, assay description, preprocessing history, and data lineage. & Noise, confounding, batch effects, missing metadata, and preprocessing bias. \\ \textbf{Software and statistical outputs} & Supports intermediate decisions, diagnostics, repair, and downstream interpretation. & Tool and version, inputs, parameters, execution logs, intermediate artifacts, and output lineage. & Tool assumptions, invalid inputs, parameter sensitivity, parsing errors, and silent failures. \\ \rowcolor{gray!10} \textbf{Scientific-model outputs} & Supports prediction, simulation, representation, ranking, or design. & Model and version, input representation, inference settings, calibration status, and output scores. & Distribution shift, miscalibration, model bias, instability, and limited interpretability. \\ \textbf{Experimental or clinical observations} & Provides empirical context, corroboration, or feedback for computational claims. & Protocol, sample or cohort definition, endpoint, assay conditions, provenance, and relation to the evaluated claim. & Cost, limited scale, protocol dependence, measurement noise, and restricted generalizability. \\ \bottomrule \end{tabular} \end{table*} 

\paragraph{Literature and structured biological knowledge.}

The scientific literature supports method selection, hypothesis generation, and biological interpretation, while databases, ontologies, pathway resources, and knowledge graphs support entity normalization and relationship retrieval. These sources can improve factual grounding and attribution, but their availability does not make a claim self-validating. Their evidentiary value depends on the quality of the source, the publication or database version, the mapping of the identifier, the retrieval procedure and the relevance to the biological context evaluated \cite{liu2025improving,amugongo2025retrieval,montori2000publication, gene2026gene,milacic2024reactome,uniprot2025uniprot}. Systems should therefore preserve citations, retrieved passages, database versions, identifiers, and query records while distinguishing established findings from extrapolation and newly generated hypotheses. 

\paragraph{Biological measurements and computational outputs.} 

Biological measurements provide dataset-specific evidence through sequencing reads, expression matrices, variants, spatial coordinates, protein-abundance measurements, images, perturbation readouts, or clinical variables. Their interpretation depends on associated metadata, including sample identity, assay type, tissue, treatment, batch, cohort definition, and experimental design. Empirical origin does not guarantee reliability: noise, confounding, batch effects, missing metadata, limited sample size, and preprocessing choices may substantially affect downstream inference \cite{yu2024assessing,luecken2019current,mao2024spatialqc, salim2025spanorm}. Software and statistical outputs provide derived computational evidence, including quality-control reports, alignments, fitted models, differential-expression results, enrichment analyses, rankings, plots, and execution logs. These artifacts often connect one workflow step to the next, but inherit the assumptions and failure modes of the tools that produced them. Their interpretation therefore requires identifiable inputs, software versions, parameters, execution status, parsing procedures, and artifact lineage. Computational outputs should be treated as provenance-bearing evidence objects rather than automatically correct observations \cite{khan2019sharing,galaxy2024galaxy,di2017nextflow}. Whether the preserved artifacts are sufficient to establish replayability or scientific validity is determined separately under Validation. 

\paragraph{Scientific-model outputs.} 

Scientific models contribute to predictions, representations, simulations, rankings, and design candidates. Examples include structure predictors, protein language models, single-cell foundation models, variant-effect predictors, docking systems, and generative molecular models \cite{jumper2021highly,lin2023evolutionary,cui2024scgpt, hao2024large,theodoris2023transfer,cheng2023accurate}. Such outputs can support candidate prioritization and hypothesis generation but should not be treated as equivalent to biological observations. The systems should record the model and version, the input representation, the inference settings, the confidence or calibration information, and the status of each output as a prediction. Additional computational or empirical testing belongs to the Validation dimension. 

\paragraph{Experimental or clinical observations.} 

Experimental and clinical observations include perturbation measurements, binding or functional assays, clinical outcomes, and other empirical readouts available to the workflow. These observations may serve as prior evidence, retrospective corroboration, or feedback for subsequent planning. Their presence does not, however, automatically establish prospective Validation. Previously generated or published observations used by a workflow are coded as \textbf{E6}. By contrast, prospective testing of an analysis, recommendation, hypothesis, or design generated by the evaluated workflow contributes to \textbf{V4} and qualifier \textbf{P}, provided that the test is aligned with the principal claim. Qualifier \textbf{C} is assigned only when the newly generated empirical result is returned to the system and changes a subsequent computational or experimental action. This distinction separates access to empirical evidence from prospective testing and genuine design--test--refine feedback.

\subsection{Validation: Reported Assurance for Workflow Outputs and Claims} \label{subsec:validation-maturity} 

The Validation dimension characterizes the assurance established for the outputs and scientific claims of an evaluated workflow. It is distinct from Function, which records demonstrated workflow operations, and Evidence, which records the sources supporting workflow actions and interpretations. Validation instead asks whether the workflow has been shown to execute, can be replayed, has undergone task-appropriate scientific evaluation, or has received prospective empirical support. This distinction is necessary because successful execution does not establish scientific correctness. A workflow may run without error while using inappropriate inputs, normalization procedures, sample groupings, parameters, statistical assumptions, or biological interpretations. Similarly, access to literature, databases, model predictions, or experimental observations does not necessarily validate the current workflow or claim \cite{ma2024agentboard,mitchener2025bixbench,guo2026promptbio,wang2026open}. We define five Validation stages, denoted \textbf{V0--V4}. These stages are cumulative assurance gates for a specific evaluated use case: \textbf{V1} requires demonstrated execution; \textbf{V2} additionally requires sufficient information for replay; \textbf{V3} requires task-appropriate scientific evaluation; and \textbf{V4} requires prospective empirical testing of an agent-generated output aligned with the principal claim. The stages are not global certifications of entire platforms, and the same system may satisfy different stages across tasks. Because a single stage cannot represent every form of support, Validation qualifiers are recorded separately: \textbf{B} for quantitative benchmark or relevant baseline comparison, \textbf{H} for protocol-based human or expert assessment, \textbf{S} for statistical diagnostics, calibration, or uncertainty analysis, \textbf{R} for robustness, sensitivity, ablation, or failure analysis, \textbf{X} for external or independent validation, \textbf{P} for prospective empirical testing, and \textbf{C} for closed-loop empirical refinement. V-stages and qualifiers apply to evaluated use cases and should be interpreted with the corresponding Function and Evidence profiles. 

\begin{table*}[!t] \centering \footnotesize \caption{\textbf{Validation dimension of the FEV framework.} V-stages represent cumulative assurance gates for a specific evaluated use case, not global maturity rankings of entire platforms.} \label{tab:validation_maturity_taxonomy} \setlength{\tabcolsep}{3pt} \renewcommand{\arraystretch}{1.20} \begin{tabular}{P{0.9cm} P{3.6cm} P{6.5cm} P{6.5cm}} \toprule \rowcolor{gray!30} \textbf{Stage} & \textbf{Validation stage} & \textbf{Minimum operational requirement} & \textbf{Typical supporting evidence} \\ \midrule \rowcolor{gray!10} \textbf{V0} & Illustrative output & The system produces explanations, hypotheses, recommendations, or workflow plans without demonstrated reliable execution. & Prompt outputs, qualitative examples, illustrative plans, screenshots, or informal expert impressions. \\ \textbf{V1} & Demonstrated execution & The system successfully invokes a tool, executes code, completes a workflow component, or generates an expected computational artifact. & Successful tool calls, completed scripts, example runs, execution outputs, or case demonstrations. \\ \rowcolor{gray!10} \textbf{V2} & Replayable computation & The system satisfies V1 and reports sufficient information to replay the evaluated computation, including identifiable inputs, parameters, dependencies, intermediate artifacts, and execution traces. & Scripts, notebooks, workflow definitions, versioned environments, parameter records, logs, intermediate files, and replay checks. \\ \textbf{V3} & Scientifically evaluated computation & The system satisfies V2 and evaluates workflow outputs using task-appropriate scientific checks beyond successful execution. & Relevant benchmarks or baselines, statistical diagnostics, robustness analyses, external datasets, uncertainty assessment, biological consistency checks, or protocol-based expert evaluation. \\ \rowcolor{gray!10} \textbf{V4} & Prospective empirical evaluation & The system satisfies V3 and prospectively tests an agent-generated output using empirical evidence appropriate to the principal scientific claim. & Perturbation experiments, binding or functional assays, prospective preclinical or clinical evaluation, or design--test--refine cycles. \\ \bottomrule \end{tabular} \end{table*}

\paragraph{Illustrative output and demonstrated execution.} 
At \textbf{V0}, a system produces explanations, hypotheses, recommendations, or workflow plans without demonstrated reliable execution. Such outputs may support ideation but remain proposals rather than executed computational artifacts. Qualitative examples, screenshots, and informal expert impressions do not establish execution or scientific reliability. At \textbf{V1}, at least one tool-mediated operation or workflow component is successfully demonstrated, such as a database query, tool call, executed script, notebook run, or expected computational artifact. V1 establishes operational feasibility beyond text generation, but not replayability or scientific appropriateness. Benchmark performance may be recorded as a qualifier, but does not establish V2 without sufficient information to replay the evaluated computation. 

\paragraph{Replayable computation.} 

At \textbf{V2}, an informed user should be able to reconstruct or replay the evaluated analytical trajectory. Relevant information includes identifiable inputs, data and metadata assumptions, parameters, software or model versions, dependencies, executable workflow definitions, intermediate artifacts, and execution traces. Containers and complete environment specifications provide strong support but are not the only acceptable forms of replayability. Replayability supports inspection, but does not establish scientific correctness. A replayable workflow may still apply an inappropriate method, omit relevant covariates, use unstable parameters, or over-interpret weak biological signals. Therefore, V2 establishes reconstructability, while V3 requires evaluation of scientifically meaningful properties of the workflow or claim. 

\paragraph{Scientifically evaluated computation.} 

At \textbf{V3}, the system satisfies V2-level replayability and evaluates its outputs using scientific checks appropriate to the task and claim. These checks may include relevant baselines, statistical diagnostics, uncertainty analysis, robustness or sensitivity testing, external datasets, biological consistency assessment, or expert evaluation under a defined protocol. V3 should not be assigned solely because a publication reports an accuracy value, qualitative demonstration, informal expert impression, or agreement with existing biological knowledge. The evaluation must test a meaningful property of the workflow or claim beyond successful execution. Because different forms of assessment provide non-equivalent assurance, qualifiers should be reported explicitly. For example, \textbf{V3 [B,R]} denotes benchmark and robustness support, whereas \textbf{V3 [B,H,X]} additionally records protocol-based expert assessment and external validation. 

\paragraph{Prospective empirical evaluation.} 
At \textbf{V4}, an output generated by the evaluated workflow is tested prospectively using empirical evidence aligned with its principal claim. Examples include testing a proposed guide RNA through gene editing, evaluating a designed binder in a binding assay, or assessing a prioritized intervention in a prospective preclinical or clinical study. Experimental data do not automatically establish V4. Previously generated measurements used as workflow inputs belong to the Evidence profile, while evaluation on archived experimental datasets remains retrospective computational Validation. V4 requires a direct relationship between the workflow-generated output and its prospective empirical test. The scope of V4 is claim-specific. For example, confirmation of a designed peptide's secondary structure does not establish binding, biological function, developability, or therapeutic utility. V4 therefore records empirical support for the evaluated claim rather than global certification of the platform. Prospective testing is also distinct from closed-loop refinement. A one-time empirical test supports the qualifier \textbf{P}. Qualifier \textbf{C} applies only when the resulting observation is returned to the system and changes a subsequent computational or experimental action. 

\paragraph{Validation qualifiers.}
The V-stage identifies the highest cumulative assurance gate satisfied by the evaluated use case, whereas qualifiers preserve the reported forms of support. Qualifiers are descriptive and non-ordinal, and they do not override unmet stage requirements. A system cannot satisfy V3 through benchmark or expert assessment alone if the evaluated computation does not meet V2 replayability. Validation therefore distinguishes whether an operation was demonstrated, the computation can be replayed, its outputs were scientifically evaluated, and a workflow-generated claim was prospectively tested. Function profiles, Evidence profiles, V-stages, and qualifiers should be interpreted together rather than collapsed into a single system score.

\section{Cross-System Synthesis through the FEV Framework} \label{sec:mapping}

\begin{figure*}[!t] \centering 
\noindent \fevpanelwide{a}{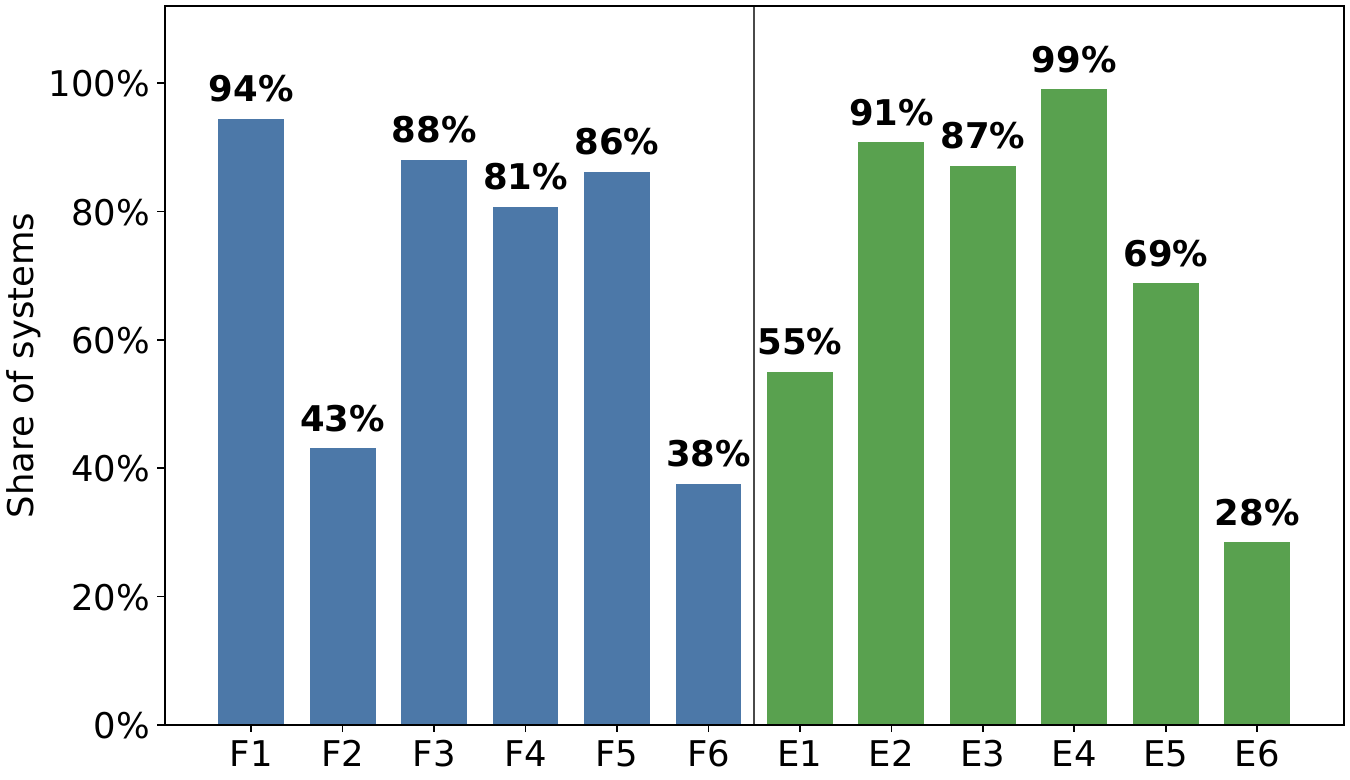}%
\hfill%
\fevpanelwide{b}{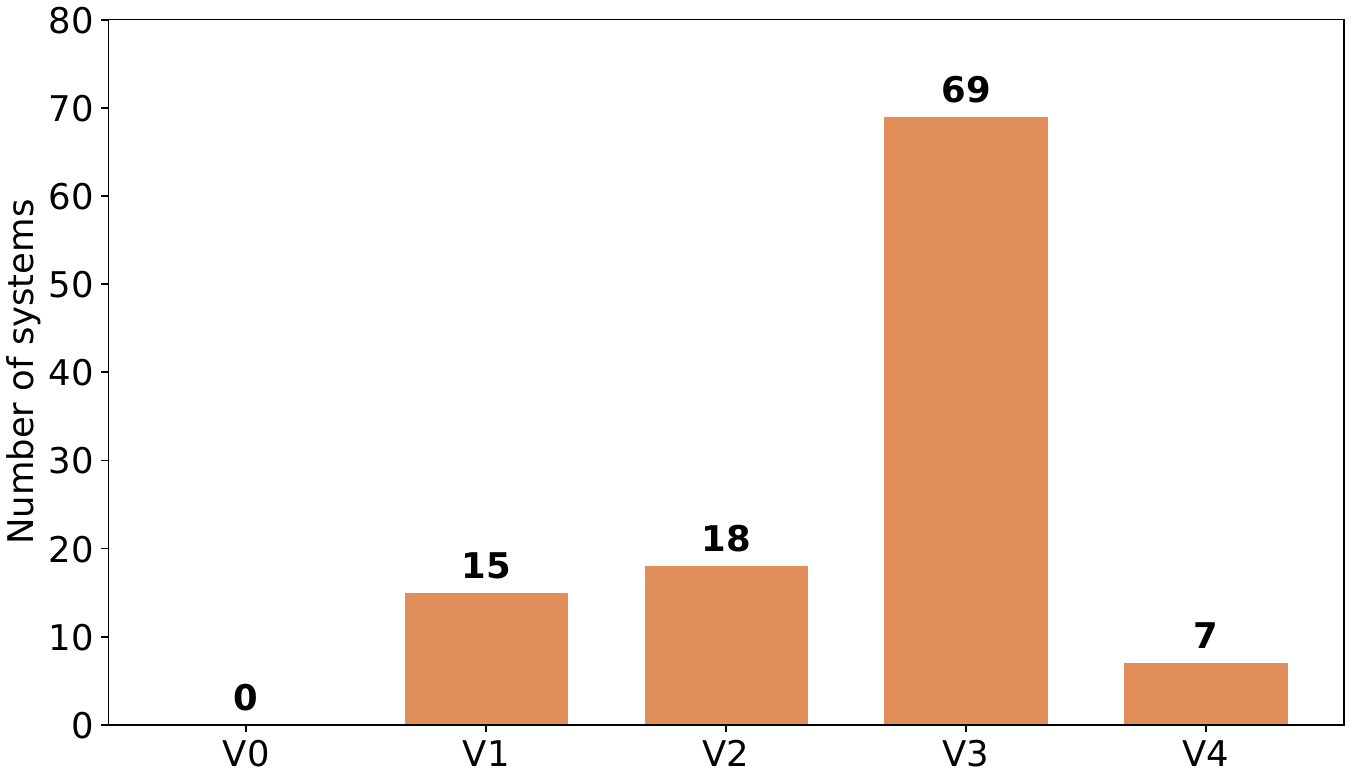} \vspace{0.35em} 
\noindent \fevpanelwide{c}{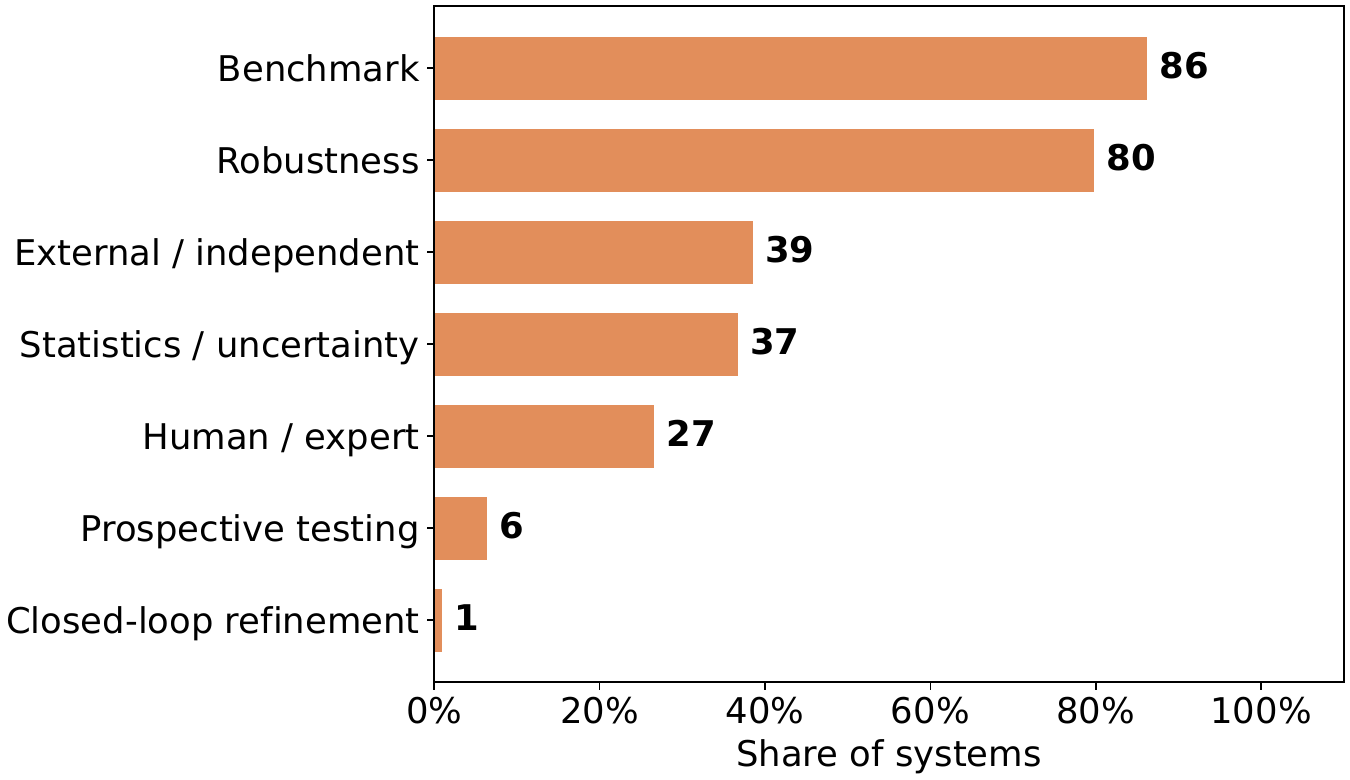}%
\hfill%
\fevpanelwide{d}{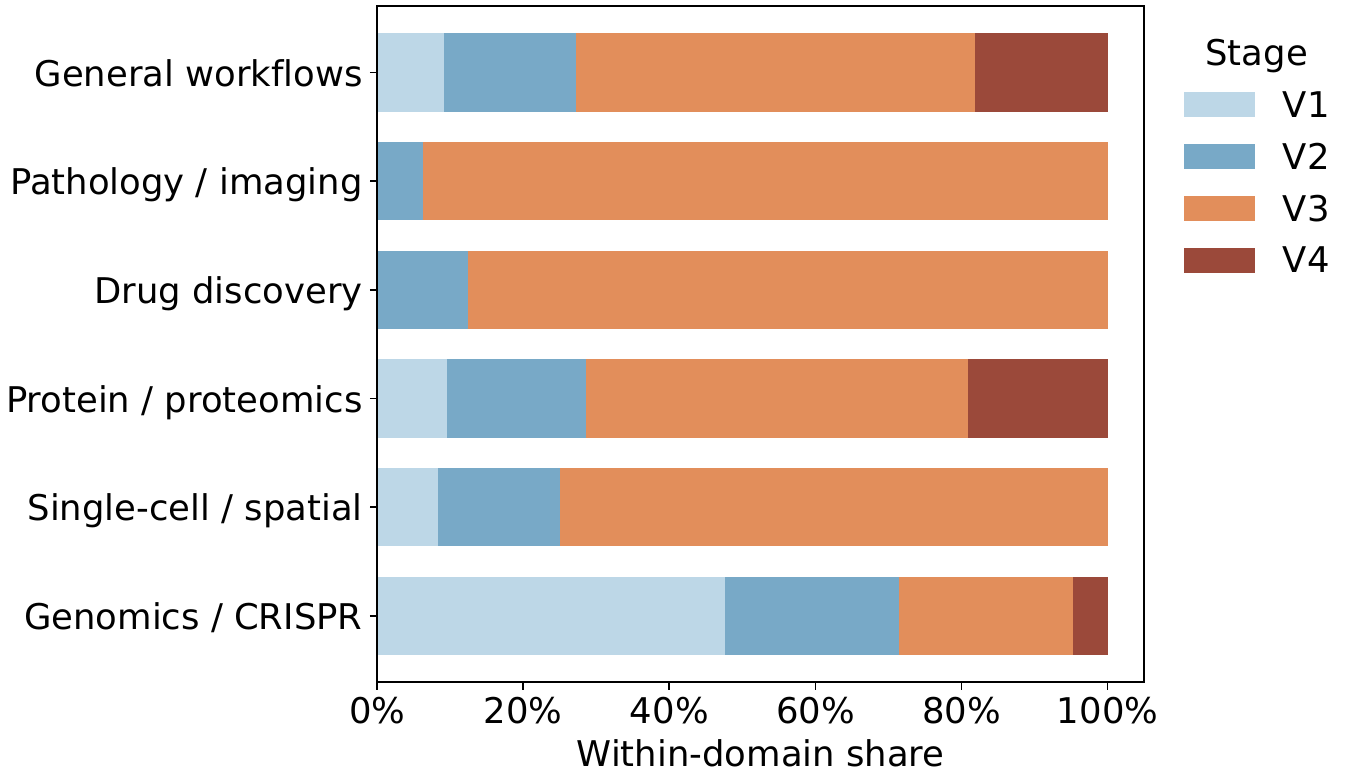} \vspace{0.25em} \caption{ \textbf{Cross-domain quantitative synthesis of the Function--Evidence--Validation landscape.} \textbf{(a)} Overall prevalence of Function and Evidence dimensions across the mapped systems. \textbf{(b)} Distribution of reported Validation stages; V0 is retained as part of the complete scale although no mapped system was assigned to this stage. \textbf{(c)} Prevalence of Validation qualifiers, including benchmark comparison, expert assessment, statistical analysis, robustness testing, external evaluation, prospective testing, and closed-loop refinement. \textbf{(d)} Within-domain composition of reported Validation stages. Counts and percentages were derived from the system-level tables in Supplementary Section S2 and describe the structured review sample rather than the complete literature. } \label{fig:fev_overall_synthesis} \end{figure*}


\begin{figure*}[!t] \centering 
\begin{minipage}[t]{0.325\textwidth} \centering \textbf{(a)}\\ \includegraphics[width=\linewidth] {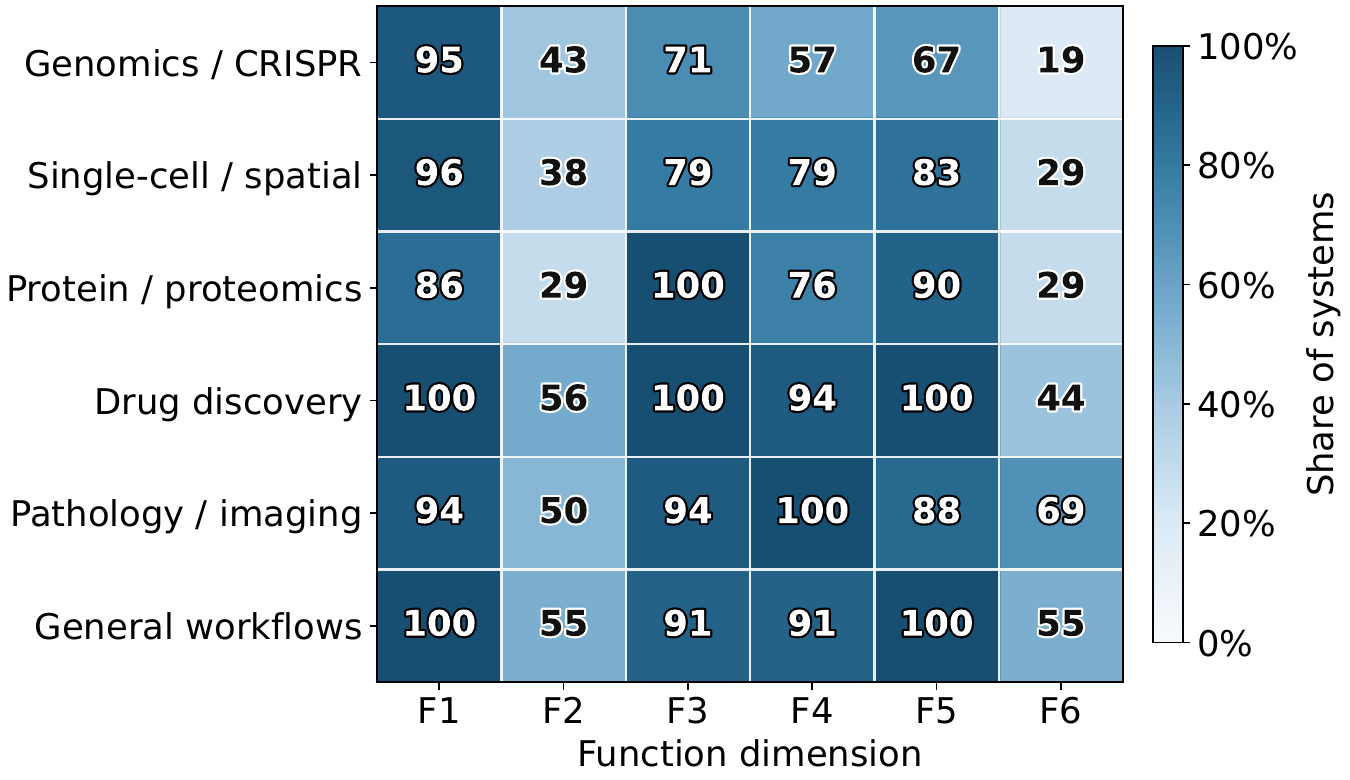} \end{minipage}%
\hspace{0.008\textwidth}
\begin{minipage}[t]{0.325\textwidth} \centering \textbf{(b)}\\ \includegraphics[width=\linewidth] {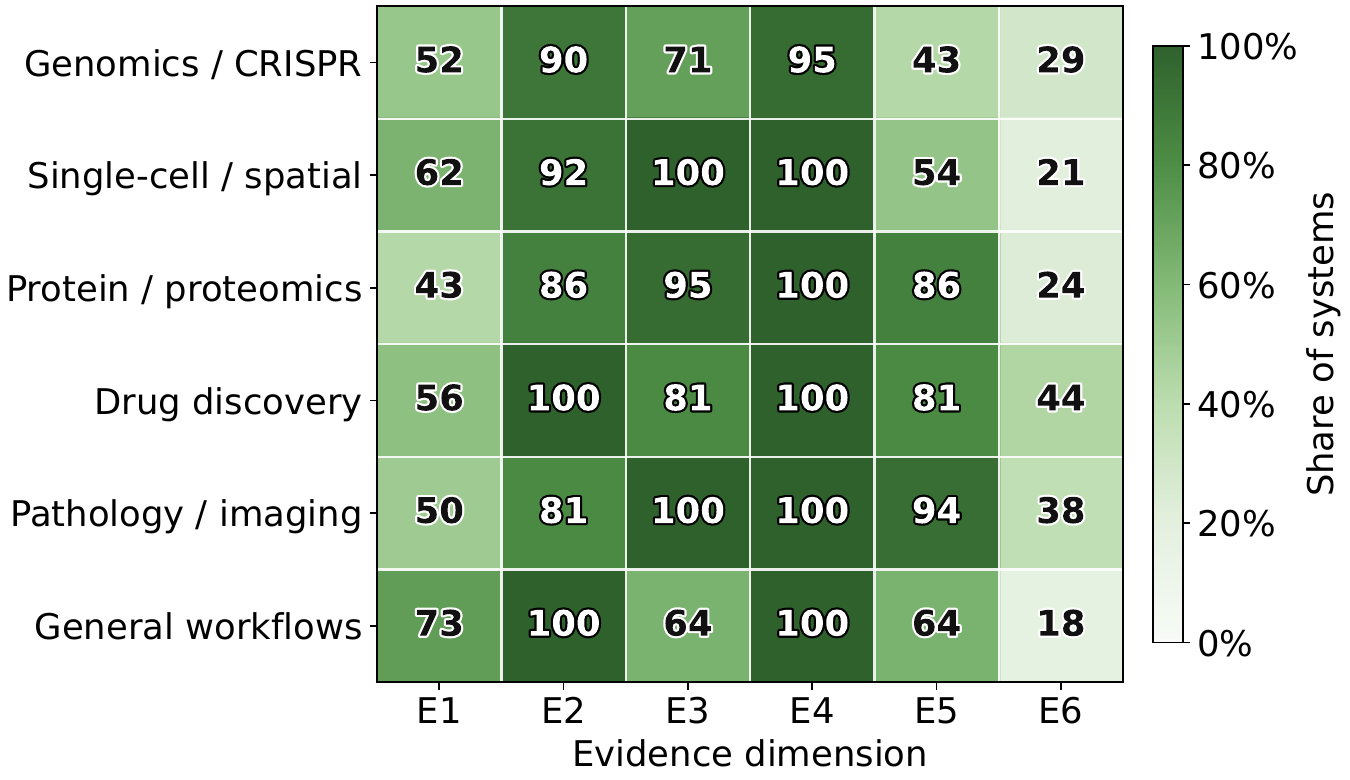} \end{minipage}%
\hspace{0.008\textwidth}
\begin{minipage}[t]{0.325\textwidth} \centering \textbf{(c)}\\ \includegraphics[width=\linewidth] {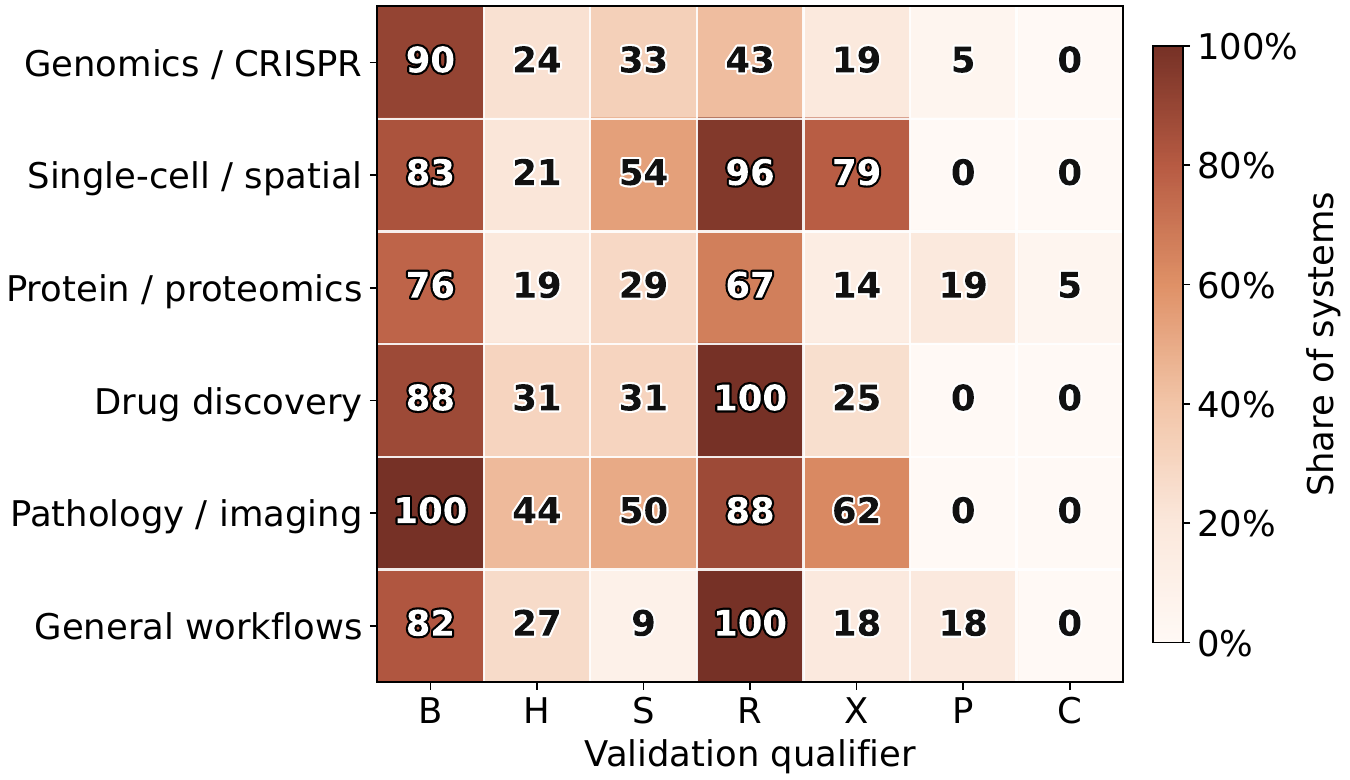} \end{minipage} \vspace{0.25em} \caption{ \textbf{Cross-domain comparison of Function, Evidence, and Validation profiles.} \textbf{(a)} Within-domain prevalence of Function dimensions F1--F6. \textbf{(b)} Within-domain prevalence of Evidence dimensions E1--E6. \textbf{(c)} Within-domain prevalence of Validation qualifiers B, H, S, R, X, P, and C. Percentages were calculated within each biological domain from the system-level tables in Supplementary Section S2. The panels describe differences in the composition of the mapped sample and should not be interpreted as rankings of fields or prevalence estimates for the complete literature. } \label{fig:fev_domain_comparison} \end{figure*}

\begin{figure*}[!t] \centering \noindent \fevpanelwide{a}{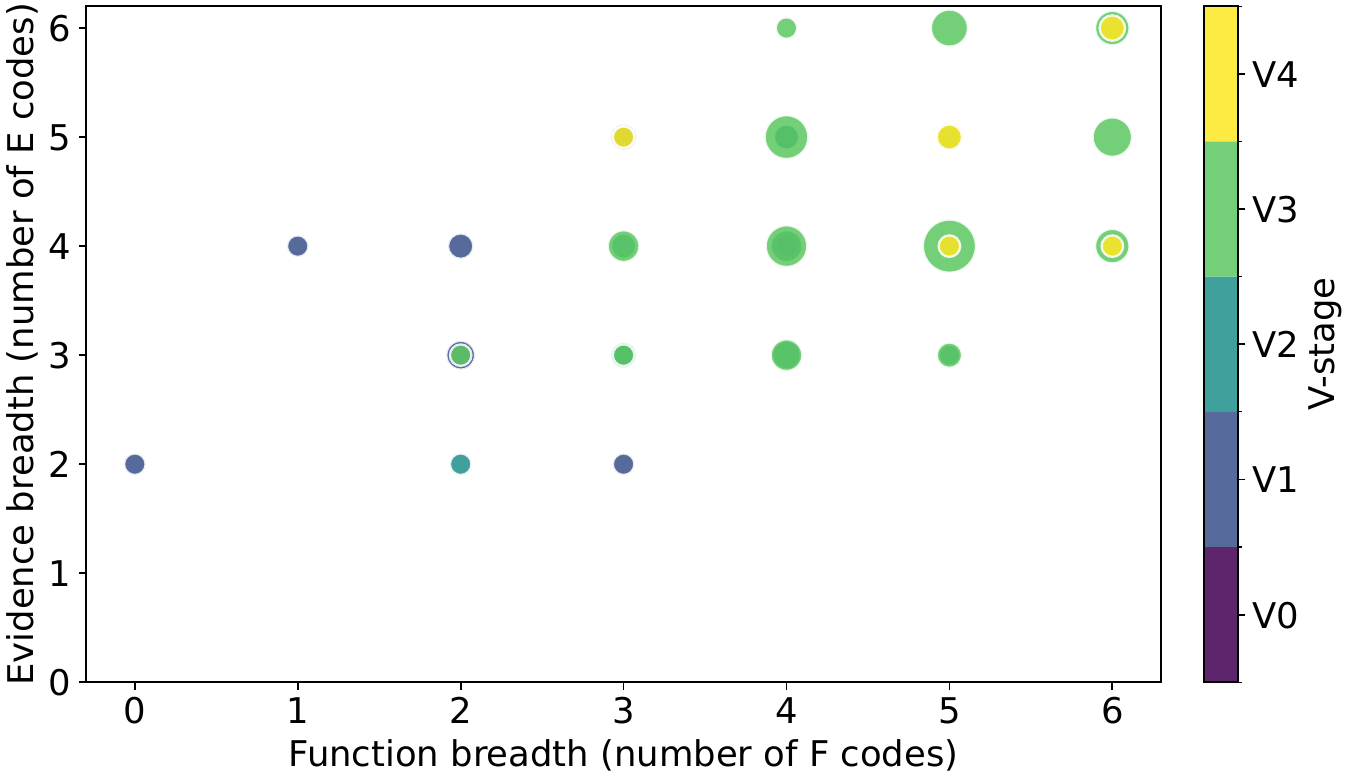}%
\hfill%
\fevpanelwide{b}{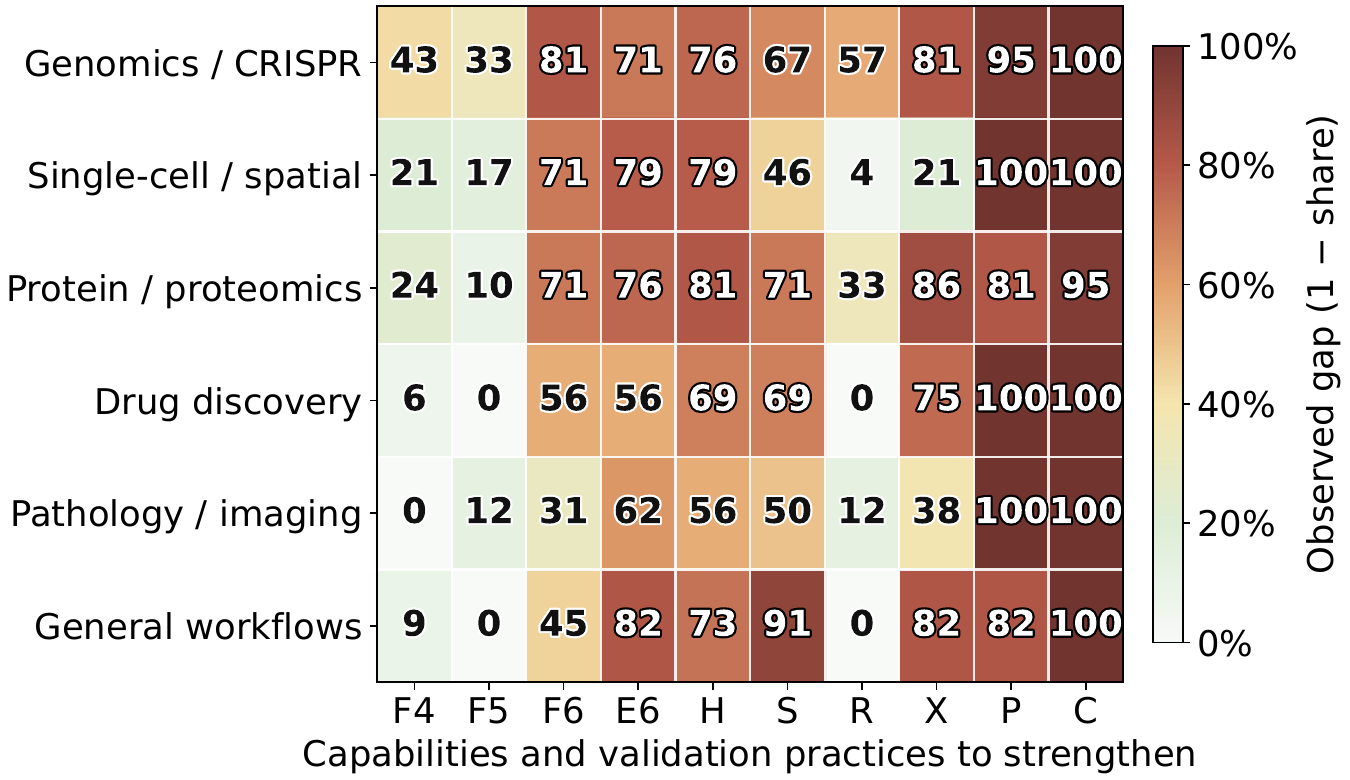} \vspace{0.25em} \caption{ \textbf{Capability breadth, reported Validation stage, and research gaps in agentic bioinformatics.} \textbf{(a)} Relationship among Function breadth, Evidence breadth, and reported Validation stage; marker size indicates the number of systems sharing each FEV profile. \textbf{(b)} Cross-domain gaps in workflow capabilities and Validation practices, calculated as one minus the within-domain share reporting each dimension; darker cells indicate larger observed gaps. The analysis is derived from the system-level tables in Supplementary Section S2 and highlights verification, prospective testing, and closed-loop refinement as priorities for future agentic bioinformatics systems. } \label{fig:fev_capability_roadmap} \end{figure*}

\paragraph{Overall FEV landscape.} 

Figure~\ref{fig:fev_overall_synthesis} summarizes the 109 mapped system entries. Planning, tool-mediated execution, workflow-state maintenance, and self-evaluation or repair are widely represented, while role specialization and explicit verification or escalation are less common. Evidence profiles are dominated by software outputs, structured biological knowledge, and retrospective biological data; experimental or clinical observations are comparatively rare. Most systems are classified as V3, while only seven reach V4 and none are assigned V0. Benchmark comparison and robustness analysis are the dominant Validation qualifiers, whereas prospective testing and closed-loop refinement remain exceptional. The mapped landscape is therefore driven primarily by evaluated computation rather than prospective empirical closure. Domain-level differences reflect task definition, benchmark availability, experimental cost, and claim scope rather than global differences in scientific maturity.

\paragraph{Cross-domain differences.} 

Figure~\ref{fig:fev_domain_comparison} compares the within-domain composition of Function, Evidence, and Validation support. Planning is common across all domains, whereas coordination, workflow-state capture, and verification are more unevenly distributed. Tool-mediated execution and trace capture are especially prominent in pathology, drug discovery, and general workflow automation but less consistently reported in genomics. Explicit verification or escalation remains uncommon in genomics, protein engineering, and single-cell or spatial analysis, indicating that neither multi-agent organization nor formal verification is universal among systems described as agentic or agent-adjacent. 

Evidence profiles largely reflect data modality. Biological measurements are prominent in single-cell, spatial, and imaging workflows, whereas scientific-model outputs are especially common in pathology, protein engineering, and drug discovery. Software and statistical outputs are nearly universal, but experimental or clinical observations remain the least frequent Evidence category across domains. 

Benchmark comparison and robustness analysis are the most widespread forms of Validation support. External evaluation is concentrated in single-cell, spatial, and pathology systems, while statistical analysis and protocol-based expert assessment remain unevenly reported. Prospective testing is concentrated in protein engineering, with fewer examples in genomics and general biomedical research; closed-loop empirical refinement appears in a single mapped framework.

These domain profiles describe different forms of reported assurance rather than rankings of scientific maturity. A V3 annotation workflow and a V3 therapeutic-design workflow satisfy the same cumulative gate but may rely on different benchmarks, statistical checks, expert assessments, and external datasets. Validation must therefore be interpreted relative to the intended use and principal scientific claim.

\paragraph{Capability breadth and priorities for future systems.} 

Figure~\ref{fig:fev_capability_roadmap} examines the relationship between Function and Evidence breadth, reported Validation, and the remaining cross-domain gaps. 

As shown in panel a, most systems implement multiple Function and Evidence dimensions, yet the majority remain at V3. Broader operational scope therefore does not imply stronger scientific assurance: additional agents, tools, databases, memory modules, or predictive models may expand a workflow without establishing replayability, verification, or empirical support for its principal claim. 

Panel b reports an observed gap score, defined as one minus the within-domain share reporting each capability or Validation practice. The largest recurring gaps are related to prospective empirical testing and closed-loop refinement, while verification, expert assessment, uncertainty analysis, and external validation remain unevenly reported. These gaps are descriptive rather than universal requirements and should be interpreted in relation to the scope, intended use, and claim of the workflow. 

Progress should therefore be assessed by workflow accountability rather than by the number of agents or tools. Priorities include inspectable shared state, provenance-preserving handoffs, explicit verification gates, calibrated uncertainty, external evaluation, and claim-aligned prospective testing. Where experimentation is feasible, empirical results should inform subsequent candidate selection, experimental design, or analysis, enabling scientifically accountable design--test--refine cycles. 

The reported proportions describe the structured review sample rather than the complete literature and depend on publication-level reporting. Unreported or insufficiently documented capabilities were not coded. System-level assignments and supporting details are provided in Supplementary Tables S1--S28.

\section{Evaluation, Trustworthiness, and Benchmarking} \label{sec:evaluation} 

Evaluation of agentic bioinformatics must extend beyond final-answer quality because biological conclusions emerge from workflows involving data and metadata selection, method choice, parameterization, tool execution, intermediate artifacts, statistical assumptions, and interpretation. A plausible endpoint may therefore conceal an inappropriate analysis, lost provenance, failed tool call, or unsupported biological claim. Recent benchmarks increasingly assess process-level behavior, including planning, tool selection, execution, artifact generation, failure recovery, citation grounding, and trajectory completeness \cite{bragg2025astabench,chen2025scienceagentbench, mitchener2025bixbench,guo2026promptbio,wang2026open,miller2025bioml}. However, most remain measures of computational competence rather than prospective biological, clinical, or experimental validity. 

\subsection{From Answer Correctness to Workflow Assurance} 

Answer correctness measures agreement with a reference answer or expert judgment, whereas workflow assurance asks whether the actions producing that answer were appropriate, inspectable, and sufficiently tested for the claim being made. Relevant criteria include method selection, valid tool inputs and parameters, required artifacts, trace completeness, evidentiary attribution, failure detection, and uncertainty reporting. This distinction is particularly important in long-horizon workflows, where a locally successful but inappropriate step may invalidate downstream results. 

Workflow assurance is also claim-dependent. A database lookup, raw-data analysis, disease-mechanism claim, therapeutic recommendation, and experimentally consequential design require different forms of support. Benchmark results should therefore be interpreted relative to workflow scope, risk, and intended use rather than treated as interchangeable measures of system quality.

Under FEV, evaluation should report three complementary properties: demonstrated workflow operations, the appropriateness and traceability of supporting Evidence, and the Validation established for the evaluated use case. These properties should not be collapsed into a single accuracy score. A system may complete a task using an inappropriate method, retrieve relevant evidence without preserving a replayable workflow, or perform strongly on a benchmark without external or empirical validation.

\subsection{Benchmark Design for Agentic Bioinformatics} 

The mapped resources suggest three complementary benchmark families. \emph{Outcome-oriented benchmarks} assess the correctness of an answer, prediction, annotation, design, ranking, or report. \emph{Process-oriented benchmarks} assess planning, tool selection, parameterization, execution, artifact generation, repair, citation grounding, and trajectory completeness. \emph{Domain-calibrated benchmarks} test whether these behaviours are appropriate for particular settings, including genomics, single-cell and spatial omics, protein design, drug discovery, and computational pathology. Detailed benchmark profiles are provided in Supplementary Tables~S29--S33.

Effective benchmark suites should combine these families using realistic multistep tasks with biological data, metadata constraints, software dependencies, expected intermediate artifacts, and criteria for both computational and biological outputs. Table~\ref{tab:benchmark_task_types} summarizes the principal task types.

\begin{table*}[!t] \centering \footnotesize \caption{Benchmark task types for evaluating agentic bioinformatics systems.} \label{tab:benchmark_task_types} \setlength{\tabcolsep}{0.5pt} \renewcommand{\arraystretch}{1.25} \begin{tabular}{P{3.2cm} P{5.2cm} P{5.0cm} P{4.0cm}} \toprule \rowcolor{gray!30} \textbf{Benchmark task type} & \textbf{What it evaluates} & \textbf{Example bioinformatics task} & \textbf{Failure modes captured} \\ \midrule \rowcolor{gray!10} \textbf{Workflow construction} & Ability to translate biological intent into an appropriate analysis plan. & Design a differential-expression or single-cell analysis workflow from a dataset description and research question. & Wrong method choice, missing preprocessing, ignored covariates, or unsuitable analysis design. \\ \textbf{Tool-mediated execution} & Ability to call tools, run code, and produce expected artifacts. & Execute quality control, normalization, clustering, enrichment, or variant-annotation steps. & Invalid tool calls, dependency failures, malformed inputs, or incomplete outputs. \\ \rowcolor{gray!10} \textbf{Debugging and repair} & Ability to detect and recover from workflow failures. & Fix broken scripts, missing packages, incorrect file paths, or failed tool calls. & Superficial repair, repeated failures, or silent output corruption. \\ \textbf{Evidence-grounded interpretation} & Ability to connect results to biological evidence without overclaiming. & Interpret gene-set enrichment, cell-type markers, perturbation effects, or candidate targets. & Hallucinated claims, unsupported mechanisms, or weak evidence attribution. \\ \rowcolor{gray!10} \textbf{Reproducibility challenge} & Ability to preserve sufficient information to replay and audit the workflow. & Reproduce an analysis from agent-generated scripts, logs, parameters, and outputs. & Missing provenance, untracked versions, non-deterministic results, or incomplete artifact lineage. \\ \textbf{Validation and robustness} & Ability to assess whether results are statistically and biologically reliable. & Compare against baseline pipelines, test parameter sensitivity, or assess biological consistency. & Overfitting, unstable conclusions, ignored uncertainty, or limited biological plausibility. \\ \bottomrule \end{tabular} \end{table*}

Benchmark tasks should combine deterministic grading of execution, files, parameters, and artifacts with domain-sensitive assessment of biological interpretation. They should also evaluate robustness under missing or contradictory metadata, corrupted inputs, unavailable tools, software-version changes, task reformulation, irrelevant retrieval, and conflicting evidence. A correct endpoint should not receive full credit when required tools were not invoked, evidence was fabricated, or the workflow cannot be replayed; conversely, a well-documented trajectory does not compensate for an incorrect biological conclusion. Finally, benchmark reports should state the claims that their results cannot support. Performance in question answering, tool selection, execution, or retrospective datasets does not establish prospective biological, preclinical, or clinical utility. Evaluation should therefore identify the highest Validation stage directly supported and the remaining requirements for external or empirical testing. 

\section{Challenges and Future Directions} \label{sec:challenges} 

The FEV synthesis shows that planning, retrieval, and tool-mediated execution are advancing more rapidly than replayability, formal verification, and prospective empirical testing. Therefore, progress depends less on increasing the number of agents or tools than on strengthening the accountability of workflow actions, evidence, and claims. 

\subsection{Reliable and Replayable Workflows} 

Real bioinformatics workflows contain incomplete metadata, heterogeneous files, missing dependencies, ambiguous objectives, and confounded study designs. Future systems should validate inputs and assumptions, detect underspecified tasks, recover from execution failures, and request clarification or additional review when available evidence is insufficient. Robustness should be tested under changes in metadata, task formulation, tool availability, software versions, and biological domains rather than only on curated demonstrations. Systems should also preserve the information required to inspect and replay their workflows, including dataset identifiers, metadata, analytical assumptions, tool calls, parameters, software and model versions, generated code, intermediate artifacts, failures, repairs, and evidence links. Agentic layers should be built on workflow engines, notebooks, containers, provenance standards, and institutional infrastructure rather than replacing them. Standardized interfaces for tools, artifacts, and evidence would improve portability between models and laboratories. 

\subsection{Claim-Calibrated Validation and Empirical Closure} 

Validation requirements should be proportional to the claim. Database retrieval, raw-data analysis, disease-mechanism inference, therapeutic recommendation, and protein design require different forms of assurance. Computational assessment may include relevant baselines, statistical diagnostics, uncertainty analysis, robustness testing, external datasets, biological consistency checks, and protocol-based expert review. Claims about biological function or intervention also require prospective testing aligned with the principal claim. Prospective testing should remain distinct from closed-loop refinement. A one-time assay evaluates an agent-generated output, whereas a closed loop returns the empirical result to the system and changes the next computational or experimental action. Greater autonomy is not inherently preferable: requesting clarification, refusing unsupported inference, or recommending additional validation may represent more reliable scientific behavior. 

\subsection{Governance, Oversight, and Reporting Standards} 

The appropriate degree of autonomy should depend on the risk of workflow. Exploratory analyses may permit broader automation, while patient-derived data, clinical interpretation, therapeutic reasoning, gene editing, and experimental design require stronger access controls, approval gates, audit logs, privacy protections, and expert oversight. Human involvement should be operationally defined by specifying which decisions require review and how the intervention changes the workflow. A minimum reporting standard should document workflow scope, models and tools, data and evidence sources, parameters and environments, intermediate artifacts, failures and repairs, approval points, reported Validation stage, and the specific forms of computational or empirical support. The most valuable agentic bioinformatics systems will therefore not necessarily be those with the greatest autonomy, but those that accelerate scientific work while preserving replayability, evidentiary traceability, uncertainty, appropriate oversight, and claim-calibrated Validation.

\section{Conclusion} \label{sec:conclusion} 

Agentic bioinformatics is extending biological computation from conversational assistance toward adaptive workflow control. Its scientific value lies not only in making analytical tools accessible through natural language, but in translating biological objectives into workflows whose actions, evidence, artifacts, and limitations can be inspected. The Function--Evidence--Validation (FEV) framework separates three properties that should not be conflated: the workflow operations a system demonstrates, the traceable evidence supporting its decisions and claims, and the assurance established for its outputs. The reviewed literature shows rapid progress in planning, retrieval, and tool-mediated execution, but less consistent support for replayable computation, evidentiary provenance, robust scientific evaluation, and prospective empirical testing. The next stage of agentic bioinformatics therefore depends less on maximizing autonomy than on strengthening scientific accountability. Reliable systems should preserve inspectable workflow trajectories, distinguish observations from predictions and hypotheses, expose uncertainty, support meaningful human oversight, and apply validation appropriate to the claims being made. Ultimately, agentic systems should be judged not only by the answers they produce but also by whether the workflows supporting those answers can be replayed, evaluated, challenged, and, where necessary, tested empirically.

\section*{Key Points} 
\begin{itemize} 
\item Agentic bioinformatics should be evaluated through inspectable workflow trajectories rather than architecture or final answers alone. 
\item The Function--Evidence--Validation framework separates demonstrated workflow operations, traceable evidentiary support, and use-case-specific scientific assurance. 
\item Among 109 system entries, planning and tool-mediated execution are more common than explicit verification, prospective testing, and closed-loop refinement. 
\item Replayability, provenance-preserving evidence linkage, calibrated validation, and expert escalation are priorities for scientifically accountable agentic workflows. 
\end{itemize} 

\section*{Author Contributions} Phuc Pham: Conceptualization, Methodology, Investigation, Data curation, Formal analysis, Visualization, Writing---original draft, Writing---review and editing. Truong Son Hy: Conceptualization, Methodology, Supervision, Writing---review and editing. 
\section*{Funding} 
This work received no specific funding.

\section*{Conflict of Interest} 
The authors declare no competing interests. 
\section*{Data Availability} 
All data supporting this review are derived from published sources cited in the manuscript and Supplementary Material.

\newpage
\clearpage
\nocite{*}
\bibliographystyle{oup-plain}
\bibliography{ref}

\newpage
\clearpage

\section*{Supplementary Material}
\input{Supplementary}

\end{document}